\documentclass{article}

\PassOptionsToPackage{numbers, compress}{natbib}

\usepackage[preprint]{neurips_2024}

\usepackage[table]{xcolor} 
\usepackage[utf8]{inputenc} 
\usepackage[T1]{fontenc}    
\usepackage{hyperref}       
\usepackage{url}            
\usepackage{booktabs}       
\usepackage{amsfonts}       
\usepackage{nicefrac}       
\usepackage{microtype}      
\usepackage{xcolor, graphicx} 
\usepackage{amsmath}
\usepackage{subcaption}
\usepackage{makecell}
\usepackage{algorithm}
\usepackage{algpseudocode}
\usepackage{multirow}
\usepackage{subcaption}
\usepackage{graphicx}
\usepackage{wrapfig}

\title{Redefining the Down-Sampling Scheme of U-Net for Precision Biomedical Image Segmentation}

\author{%
  Mingjie Li, Yizheng Chen, Md Tauhidul Islam, Lei Xing\\
  Department of Radiation Oncology\\
  Stanford University, CA, USA\\
  \texttt{\{lmj695,chenyz,tauhid,lei\}@stanford.edu} \\
}

\begin{document}

\maketitle

\begin{abstract}
U-Net architectures have been instrumental in advancing biomedical image segmentation (BIS) but often struggle with capturing long-range information. One reason is the conventional down-sampling techniques that prioritize computational efficiency at the expense of information retention. This paper introduces a simple but effective strategy, we call it Stair Pooling, which moderates the pace of down-sampling and reduces information loss by leveraging a sequence of concatenated small and narrow pooling operations in varied orientations. Specifically, our method modifies the reduction in dimensionality within each 2D pooling step from $\frac{1}{4}$ to $\frac{1}{2}$. This approach can also be adapted for 3D pooling to preserve even more information. Such preservation aids the U-Net in more effectively reconstructing spatial details during the up-sampling phase, thereby enhancing its ability to capture long-range information and improving segmentation accuracy. Extensive experiments on three BIS benchmarks demonstrate that the proposed Stair Pooling can increase both 2D and 3D U-Net performance by an average of 3.8\% in Dice scores. Moreover, we leverage the transfer entropy to select the optimal down-sampling paths and quantitatively show how the proposed Stair Pooling reduces the information loss.
\end{abstract}

\section{Introduction}

Biomedical image segmentation (BIS)~\cite{BISsurvey,islamaj2024overview} is a critical task in diagnostic imaging, facilitating the accurate delineation of anatomical structures and pathological regions. The U-Net~\cite{unet} and a range of its variants~\cite{nnunet,zhou2018unet++,huang2020unet3+,chen2021transunet,cao2022swinu,wang2024mamba,cciccek20163d,ma2024segment,chen2025inference,weng2023mask} have significantly advanced the state-of-the-art (SOTA) in this domain, owing to their effective feature extraction and multi-scale information integration capabilities. However, one of the persistent challenges with U-Net that is often criticized is the limited efficacy in capturing long-range information relationships within medical images. On the one hand, CNN-based methods hard to learn explicit global and long-range semantic interactions~\cite{chen2021transunet}. To this end, attention mechanism-based backbones are leveraged to replace the CNN blocks and achieve excellent performances in the field of BIS. However, these models become redundant and involve significant extra computation costs. And except for more training samples to feed them. 

On the other hand, this limitation largely stems from the traditional approach to down-sampling employed in these networks, such as strided convolutions~\cite{soomro2018strided,riad2021learning} or typical pooling operations~\cite{lecun1998gradient}. Although these techniques facilitate a rapid reduction in spatial dimensions, which enlarges the receptive fields of network layers and enhances robustness to input variations, they prioritize computational efficiency at the expense of retaining critical feature details~\cite{zhao2020liftpool}. This considerable and non-invertible loss of information compromises the network's ability to capture long-range dependencies, which is especially detrimental in BIS where precise segmentation of fine structures is crucial for accurate diagnosis and treatment planning. Recently, various novel pooling techniques have been proposed to enhance the long-range semantic interaction during the down-sampling process, which can be divided into pyramid-based~\cite{du2021unet,li2023automatic} and wavelet-based~\cite{zhou2023xnet,xu2023haar,zhao2023wranet} methods. Although they leverage multi-scale features or multi-frequency information to enhance the long-range semantic information interactions, the minimum pooling receptive field is $2\times2$. It means that every pooling operation will compress four position information into one. Therefore, a natural question is raised: \textit{Can we find smaller pooling layers and slow down the current down-sampling techniques?}

Specifically, the core idea of this paper is to utilize small and narrow pooling kernels to progressively down-sample feature maps, as shown in Figure.\ref{fig:stair pool}. However, transitioning directly from high-dimensional pooling layers to a series of concatenated low-dimensional pooling layers presents two main challenges. First, it is necessary to break the linear relationship between the concatenated pooling layers, as this setup would essentially replicate the original high-dimensional pooling layer. Second, it is crucial to manage features from different down-sampling paths. For instance, consider splitting a 2D pooling layer where one path performs pooling in the vertical direction first and then in the horizontal direction, while the other path does the reverse. Splitting a 3D pooling layer would result in even more paths. Therefore, another question arises whether it is possible to identify the optimal down-sampling pathway.

In response to these challenges, this paper introduces a simple but effective modification to the down-sampling strategy used in U-Net architectures, we call it ``Stair Pooling''. Our method is designed to moderate the rate of down-sampling, thus balancing computational demands with the need for high information retention. Stair Pooling employs a series of concatenated small and narrow pooling operations, implemented in varied orientations, to control the reduction in spatial resolution more gradually. Specifically, we adjust the typical reduction in dimensionality during each 2D pooling step from $\frac{1}{4}$ to a more conservative $\frac{1}{2}$, which allows for the preservation of more critical information across the down-sampling process. Each pooling layer is followed by a convolution layer and ReLU function to interact features and break the linear relationships. Then we fuse the features from all paths together to attain the final down-sampled features. Moreover, the proposed Stair Pooling can be extended to 3D pooling operations prevalent in volumetric BIM tasks. In the end, we leverage the entropy transfer to approximate the interaction between the down-sampling layers and the final output layer and find the best down-sampling path.

To validate the effectiveness of Stair Pooling, we conducted extensive experiments across two 2D BIS benchmarks (Synapse and ACDC) and one 3D BIS benchmark (KiS23). Our results demonstrate that integrating Stair Pooling into both 2D and 3D U-Net architectures enhances their performance significantly, increasing the average Dice scores by 3.8\%. By leveraging transfer entropy, we can further remove low-information down-sampling paths and make the network simpler without performance drop and imposing additional computational burdens. These findings underscore the potential of Stair Pooling to advance the capabilities of U-Net architectures in BIS by ensuring a better balance between computational efficiency and the preservation of critical diagnostic information.

\section{Related works}
\subsection{U-Net Architectures for Biomedical Image Segmentation}

The U-Net architecture (encoder-decoder architecture~\cite{li2023dynamic,li2022mitigating}) leverages a fully convolutional network with a distinctive symmetric shape that promotes precise localization, establishing itself as a revolutionary framework in the field of BIS. Its variants further improve segmentation accuracy by enhancing the efficacy in capturing long-range semantic interactions. One category of these variants leverages the attention mechanism. Oktay \textit{et al.}\cite{oktay2022attention} proposed an attention gate integrated with CNNs to scale input features with attention coefficients by progressively suppressing irrelevant feature responses. Transformer architectures\cite{oktay2022attention} have also been incorporated to capture both long-range dependencies and local features, improving performance in models such as TransUNet~\cite{chen2021transunet}, SWinUNET~\cite{cao2022swinu}, UNETR~\cite{hatamizadeh2022unetr}, and Mamba-U~\cite{wang2024mamba}. However, with increasing parameters, these methods also escalate the demands for computational resources and training samples. Another line of research targets the improvement of traditional pooling techniques~\cite{nirthika2022pooling}. The pyramid pooling technique aims to fuse different scale features during the down-sampling process and has been utilized to enhance BIS such as retinal blood vessels~\cite{unet}, MRI prostate~\cite{li2023automatic}, and the spine~\cite{liu2021application}. In contrast, some works utilize wavelet transforms to replace the up- and down-sampling layers due to their powerful frequency and spatial representation capabilities. The key idea of wavelet pooling is to use two filters to encode low- and high-frequency information, exploring them to represent the complete semantics. Xu \textit{et al.}\cite{xu2023haar} leveraged fixed Haar Wavelet filters for this transformation and feature fusion. Zhao \textit{et al.}\cite{zhao2023wranet} retained only the low-low frequency semantics, removing high-frequency components to eliminate noise. Unlike these approaches that decompose features during each down-sampling process, Zhou \textit{et al.}\cite{zhou2023xnet} decomposed the input image into low- and high-frequency branches, conducted down-sampling, and fused them at the bottom with a fusion module. However, most wavelet pooling techniques are only suitable for specific segmentation objects\cite{zhou2023xnet}. Our method focuses on exploring another kind of feature to improve existing pooling techniques while maintaining efficiency.

\subsection{Pooling Techniques}
Pooling techniques~\cite{alsallakh2022mind,stergiou2022adapool,stergiou2021refining,zhao2020liftpool,gao2019lip,hou2020strip} aim to increase the receptive fields of convolutional neural networks (CNNs), attracting considerable attention for their efficacy in capturing long-range contextual information and reducing computational requirements, thereby enabling the construction of deeper architectures. Initial methods were based on the spatial pyramid pooling technique~\cite{he2015spatial}, which employs a set of parallel pooling operations with distinct kernel sizes at each pyramid level to capture large-range context. Strip pooling~\cite{hou2020strip} replaces square kernels with $1 \times N$ and $N \times 1$ kernels, facilitating the capture of long-range dependencies in isolated regions. Although this technique is integrated with CNN blocks, it inspires the utilization of non-square kernel sizes during the down-sampling process. Several approaches employ grid-sampling with learned weights, such as S3Pool~\cite{zhai2017s3pool}, DPP~\cite{saeedan2018detail}, and LIP~\cite{gao2019lip}. To mitigate information loss, Zhao and Snoek~\cite{zhao2020liftpool} introduced LiftDownPool and LiftUpPool, combining the Lifting Scheme and Wavelet transform. Nevertheless, the performance of these methods is dependent on sub-network structures, which constrains their applicability as computationally and memory-efficient pooling techniques. Recently, Stergiou and Poppe~\cite{stergiou2022adapool} proposed Adapool, which learns a region-specific fusion of the exponent of the Dice-Sørensen coefficient and exponential maximum pooling to achieve computational and memory efficiency. In this study, we propose the utilization of narrow max-pooling kernels for the down-sampling process. This strategy not only enhances the U-Net architecture's ability to capture long-range semantics by minimizing information loss during down-sampling but also maintains computational efficiency. We compare the proposed Stair Pooling with parts of generic pooling techniques in BIS tasks in \S\ref{pa:eff}, demonstrating our effectiveness and efficiency.

\section{Methods}
In this section, we first briefly introduce the preliminaries of the U-Net for BIS tasks. Then we propose Stair Pooling, a simple but effective strategy to replace the down-sampling techniques in the existing U-Net architectures. To quantify the information carried out by each path, we introduce the entropy and calculate the transfer entropy to select the best down-sampling path.

\begin{figure}[t]
    \centering
    \includegraphics[width=1\linewidth]{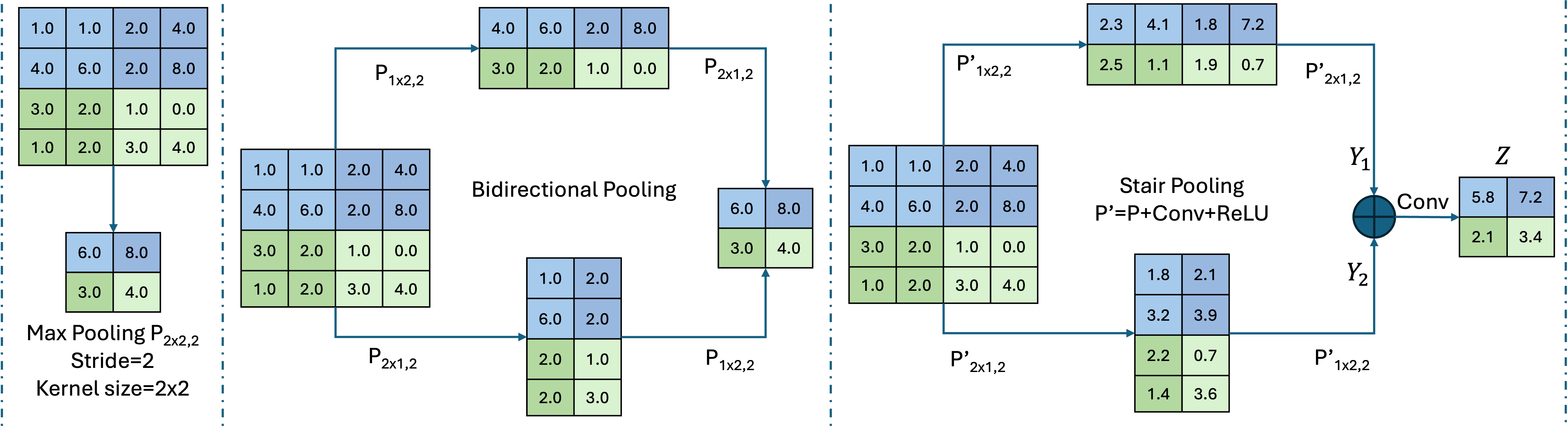}
    \caption{The overview of our proposed Stair Pooling. It splits the original max pooling layer into a series of concatenated small and narrow pooling kernels. To break the linear relationship, each pooling operation is followed by a convolutional layer and a ReLU activation.}
    \label{fig:stair pool}
\end{figure}

\subsection{Preliminaries}

The U-Net architecture is designed as a convolutional neural network optimized for tasks requiring precise localization, especially for biomedical image segmentation tasks. The network's architecture features a symmetric encoder-decoder structure. The encoder consists of repeated applications of two $3 \times 3$ convolutions (unpadded), each followed by a rectified linear unit (ReLU) and a $2 \times 2$ max pooling operation with a stride of 2 for down-sampling, where each step doubles the number of feature channels.

In contrast, the decoder path involves up-sampling the feature map followed by a $2 \times 2$ convolution ("up-convolution") that halves the number of feature channels, a concatenation with the corresponding cropped feature map from the encoder path, and two $3 \times 3$ convolutions, each followed by a ReLU. The architecture is designed to have a large number of feature channels on the up-sampling path to propagate contextual information to higher-resolution layers. The final layer is a $1 \times 1$ convolution that maps each 64-component feature vector to the number of desired classes.

The objective function commonly employed in BIS tasks combines cross-entropy loss with the Dice coefficient to balance pixel-wise classification accuracy with overlap between the predicted segmentation and the ground truth:
\begin{equation}
\mathcal{L} =   \mathcal{L}_{CE} + \mathcal{L}_{DICE} \\ = -(y \log(\hat{y}) + (1 - y) \log(1 - \hat{y})) + 1 - \frac{2 \sum y \hat{y}}{\sum y + \sum \hat{y}}
\end{equation}
where $y$ is the ground truth mask, $\hat{y}$ is the predicted mask, and $\sum$ denotes the sum over all pixels.

\subsection{Stair Pooling}

The core idea of Stair Pooling is to utilize small and narrow pooling kernels to progressively down-sample feature maps, mitigating the information loss typically associated with high-dimensional pooling operations. Unlike traditional methods that reduce dimensionality by a factor of four (e.g., $2 \times 2$ pooling), Stair Pooling reduces it more conservatively, by a factor of two, thereby preserving more critical feature information across the down-sampling process. The overview of the proposed Stair Pooling is presented in Fig.\ref{fig:stair pool}.

Instead of performing a single $2 \times 2$ pooling operation, we split it into a sequence of $1 \times 2$ and $2 \times 1$ pooling operations. For 3D pooling, the operations are similarly split into lower-dimensional components. Mathematically, for a feature map $Y$, the pooling operation can be represented as:
\begin{equation}
        Y = P_{2 \times 2}(X) \approx P_{1 \times 2}(P_{2 \times 1}(X))
\end{equation}
where $P_{m \times n}$ represents the pooling operation with a window of size $m \times n$. Each pooling operation is followed by a convolution layer and a ReLU activation function. This step ensures that the linear relationships between the concatenated pooling layers are broken, preventing redundancy and maintaining the richness of feature representation:
\begin{equation}
    Y = \text{ReLU}(\text{Conv}(P_{1 \times 2}(\text{ReLU}(\text{Conv}(P_{2 \times 1}(X))))))
\end{equation}
    
After applying the lower-dimensional pooling operations in various orientations, the features from all paths are fused together to attain the final down-sampled features. This fusion step ensures that information from different spatial dimensions is integrated, enhancing the network’s ability to capture detailed and long-range dependencies:
\begin{equation}
    Z = \text{Conv}([Y_1;Y_2;\ldots;Y_n])
\end{equation}
where $Y_i$ represents the features from each pooling path and $[;]$ refers to the concatenation operation. During each 2D pooling step, the typical reduction in dimensionality from $\frac{1}{4}$ is adjusted to a more conservative $\frac{1}{2}$. This gradual reduction allows for better retention of critical information, improving the network’s ability to handle fine structures essential for BIS tasks. Stair Pooling can be extended to 3D pooling operations prevalent in volumetric BIS tasks. By adopting this strategy, we maintain a balance between computational efficiency and the retention of essential diagnostic information.

\subsection{Transfer Entropy}

\begin{algorithm}
\caption{Stair Pooling Optimization using Transfer Entropy}
    \begin{algorithmic}[1]
        \Require A U-Net, final output feature map $X_o$, down-sampling features $Z^l$
        \Ensure Optimized down-sampling paths
        \State Initialize down-sampling paths $Y_i$, $i \in \{1, \ldots, n\}$
        \For{each down-sampling path $Y_i$}
            \State Compute the standard deviation $\sigma(X_o)$ for each channel $j$ of $X_o$
            \State Calculate the entropy $H(X_o)$ by Eq.\ref{eq: entropy}
            \State Calculate the conditional entropy $H(X_o \mid Y_i)$ of the output features with the given down-sampled features $Y_i$
            \State Calculate the Transfer Entropy $TE_{Y_i \to X_o}$ by Eq.\ref{eq: te}
        \EndFor
        \State Identify the down-sampling path $Y_{\text{opt}}$ that maximizes $TE_{Y_i \to X_o}$
        \State Optimize the Stair Pooling mechanism using the selected down-sampling path $Y_{\text{opt}}$
        \State \Return A set of optimized down-sampling paths $Y_{opt}$
        \end{algorithmic}
\label{algorithm: te}
\end{algorithm}

Here raises a natural question, is it necessary to keep all the dowm-sampling paths $Y_i$?
To optimize the down-sampling paths in Stair Pooling, we introduce Transfer Entropy (TE) to dynamically adjust the pooling paths to maximize the retention of critical information during the down-sampling process. We summarize this pipeline in Algorithm.\ref{algorithm: te}.

\noindent\textbf{Entropy Quantification} Entropy can measure the information quantity of a network. For a certain layer, the entropy can be calculated given the probability distribution of its features. However, directly measuring this distribution is challenging. Following established methods~\cite{lin2024mlp,sun2022entropy}, we use the Gaussian distribution to approximate the intermediate features in a layer. Thus, the entropy \(H(F)\) of a feature map \(F\) can be expressed as:
\begin{equation}
H(F) = -\int p(f) \log p(f) \, df, \quad f \in F.
\end{equation}
Using the Gaussian distribution approximation, this simplifies to:
\begin{equation}
H(F) = \log(\sigma) + \frac{1}{2} \log(2\pi e),
\end{equation}
where \(\sigma\) is the standard deviation of the feature set \(f \in F\). The entropy of each layer is proportional to the summation of the logarithm of the standard deviation of each feature channel:
\begin{equation}
H_\sigma(F) \propto \sum_j \log[\sigma(F_j)],\label{eq: entropy}
\end{equation}
where \(\sigma(F_j)\) calculates the standard deviation of the \(j\)-th channel of the feature set \(F\).

\noindent\textbf{Transfer Entropy Calculation} To measure the interaction between layers, we utilize TE, which quantifies the directed information transfer between two layers. Given a down-sampling path, the TE from the down-sampled features \(Y_i\) to the output features \(X_o\) is calculated as follows:
\begin{equation}
TE_{Y_i \to X_o} = H(X_o \mid Y_i) - H(X_o),\label{eq: te}
\end{equation}
where \(H(X_o)\) is the original entropy of the output features using the fused feature $Z$ and \(H(X_o \mid Y_i)\) is the conditional entropy of the output features with the given down-sampled features. By calculating the TE for each down-sampling path, we can identify the path that maximizes the information transfer to the output layer, ensuring that critical information is preserved during down-sampling. As there are only three down-sampling steps in the U-Net, we employ the exhaustive search to select the most informative down-sampling paths and optimize the Stair Pooling mechanism. Once chose the downsampling path, the fused feature $Z^l$ is equal to $Y_{opt}^l$, where $l$ refers to the $l$-th down-sampling process.

\section{Experiments}

\subsection{Settings}

To evaluate our proposed Stair Pooling, we utilized two 2D-BIS benchmarks and one 3D-BIS benchmark: the Synapse multi-organ segmentation dataset (Synapse), the Automated Cardiac Diagnosis Challenge dataset (ACDC), and the Kidney Tumor Segmentation Challenge 2023 dataset (KiTS23). 

\noindent\textbf{Synapse} The Synapse dataset comprises 30 cases with a total of 3,779 axial abdominal clinical CT images. Consistent with prior work \cite{cao2022swinu,chen2021transunet}, we utilized the official split dataset, including 18 samples for training and 12 samples for testing. Evaluation of our method on this dataset focuses on eight abdominal organs: aorta, gallbladder, spleen, left kidney, right kidney, liver, pancreas, and stomach. The average Dice-Similarity Coefficient (DSC) and average Hausdorff Distance (HD) are employed as evaluation metrics.

\begin{table}[t]
\centering
\caption{Segmentation accuracy of different methods on the Synapse multi-organ CT dataset. The highest value in each column is highlighted in bold. Gray rows refer to methods whose sizes are significantly larger than others.}\label{tab:synpase}
\resizebox{\textwidth}{!}{\begin{tabular}{lcccccccccc} 
\toprule
\multicolumn{1}{c}{Methods} & DSC↑           & HD↓            & Aorta         & Gallbladder    & Kidney(L)      & Kidney(R)      & Liver          & Pancreas       & Spleen         & Stomach         \\ 
\midrule
R50 U-Net                   & 74.68          & 36.87          & 87.74         & 63.66          & 80.6           & 78.19          & 93.74          & 56.9           & 85.87          & 74.16           \\
U-Net                       & 76.85          & 39.7           & 89.07         & 69.72          & 77.77          & 68.6           & 93.43          & 53.98          & 86.67          & 75.58           \\
R50 Att-UNet                & 75.57          & 37.66          & 55.92         & 63.91          & 79.2           & 72.71          & 93.56          & 49.37          & 87.19          & 74.95           \\
Att-UNet                    & 77.77          & 36.87          & 89.55         & 68.88          & 77.98          & 71.11          & 93.57          & 58.04          & 87.3           & 75.75           \\
\rowcolor[rgb]{ .906,  .902,  .902} R50 ViT                     & 71.29          & 32.27          & 73.73         & 55.13          & 73.52          & 72.2           & 91.51          & 45.99          & 81.99          & 73.95           \\
\rowcolor[rgb]{ .906,  .902,  .902} TransUnet                   & 77.48          & 31.69          & 87.23         & 63.13          & 81.87          & 77.02          & 94.08          & 55.86          & 85.08          & 75.62           \\
\rowcolor[rgb]{ .906,  .902,  .902} SwinUnet                    & 79.13          & \textbf{21.55} & 85.47         & 66.53          & 83.28          & 79.61          & 94.29          & 56.58          & \textbf{90.66} & 76.6            \\ 
HWT UNet                    & 77.27          & 34.16          & 88.01         & 64.09          & 81.52          & 75.2          & 94.21          & 54.12          & 87.17           & 75.75           \\
Pyramid UNet                    & 79.3          & 32.31          & 88.82         & 65.76          & 82.7          & 74.8          & \textbf{96.16}          & 57.77          & 88.8           & 73.85           \\
\midrule
SP U-Net                    & 80.45          & 31.25          & 89.26         & \textbf{71.59} & 83.26          & 76.53          & 94.77          & \textbf{67.75} & 86.65          & 73.8            \\
w. TE                        & \textbf{80.89} & 31.18          & \textbf{90.2} & 70.52          & \textbf{84.43} & \textbf{79.76} & 95.09 & 63.46          & 86.69          & \textbf{76.98}  \\
\bottomrule
\end{tabular}}
\end{table}

\noindent\textbf{ACDC} The ACDC dataset consists of MRI scans collected from various patients. For each patient's MR image, three structures are labeled: the left ventricle (LV), right ventricle (RV), and myocardium (MYO). We followed the same data split and pre-processing by \cite{chen2021transunet}, and finally got 70 training samples, 10 validation samples, and 20 testing samples. To compare with the previous works, we use the average DSC as the evaluation metric on this dataset.

\paragraph{KiT23} The KiTS23 dataset is the third iteration of the Kidney Tumor Segmentation Challenge, following editions held in 2019 and 2021. This year's challenge features an expanded training set with 489 cases and a test set comprising 110 cases. The imaging data and ground truth labels are provided in NII format with dimensions (slices, height, width). In managing class imbalance, we over-sampled the foreground representing the kidney by assigning a probability of 0.5. 

\begin{wraptable}{r}{0.45\textwidth}
\centering
\caption{Segmentation accuracy of different methods on the ACDC dataset in DSC($\%$).}\label{tab:acdc}
\resizebox{0.45\textwidth}{!}{\begin{tabular}{lcccc}
\toprule
\multicolumn{1}{c}{Methods} & DSC   & RV    & Myo   & LV     \\
\midrule
R50 U-Net                   & 87.55 & 87.1  & 80.63 & 94.92  \\
U-Net                       & 86.63 & 85.46 & 81.08 & 93.35  \\
\rowcolor[rgb]{ .906,  .902,  .902} R50 ViT                     & 87.57 & 86.07 & 81.88 & 94.75  \\
\rowcolor[rgb]{ .906,  .902,  .902} TransUnet                   & 89.71 & \textbf{88.86} & 84.53 & 95.73  \\
\rowcolor[rgb]{ .906,  .902,  .902} SwinUnet                    & 90.00   & 88.55 & \textbf{85.62} & 95.83  \\
Pyramid UNet       & 89.17 & 88.14  & 83.85 & 95.52  \\
HWT UNet           & 87.84 & 85.47 & 83.32 & 94.73  \\
\midrule
SP U-Net                    & \textbf{90.18} & 88.81 & 84.63 & 95.90  \\
w. TE                       & 90.08 & 87.74 & 84.14 & \textbf{96.08} \\
\bottomrule
\end{tabular}}
\end{wraptable}
The UNet with the proposed Stair Pooling is implemented based on Python 3.7.12 and Pytorch 1.13.1 with CUDA 11.7. We used flip and rotations as data augmentation techniques to increase training data diversity. For 2D data, input image size and patch size are set as $224\times224$ and 4, respectively. For 3D data, we standardized all images to a common voxel spacing of $1.84\times1.84\times2.35$. The network was trained using a patch size of 128×128×128. We train the network on two NVIDIA RTX A5000 GPUs with 24GB memory. During the training period, the batch size and epoch are 24 and 250 for 2D, and 2 and 100 for 3D images, respectively. The optimizer is the popular SGDD with a momentum of 0.9 and weight decay 1e-4.

\subsection{Main Results}
\label{4.2}

\begin{figure}[t]
    \centering
    \includegraphics[width=1\linewidth]{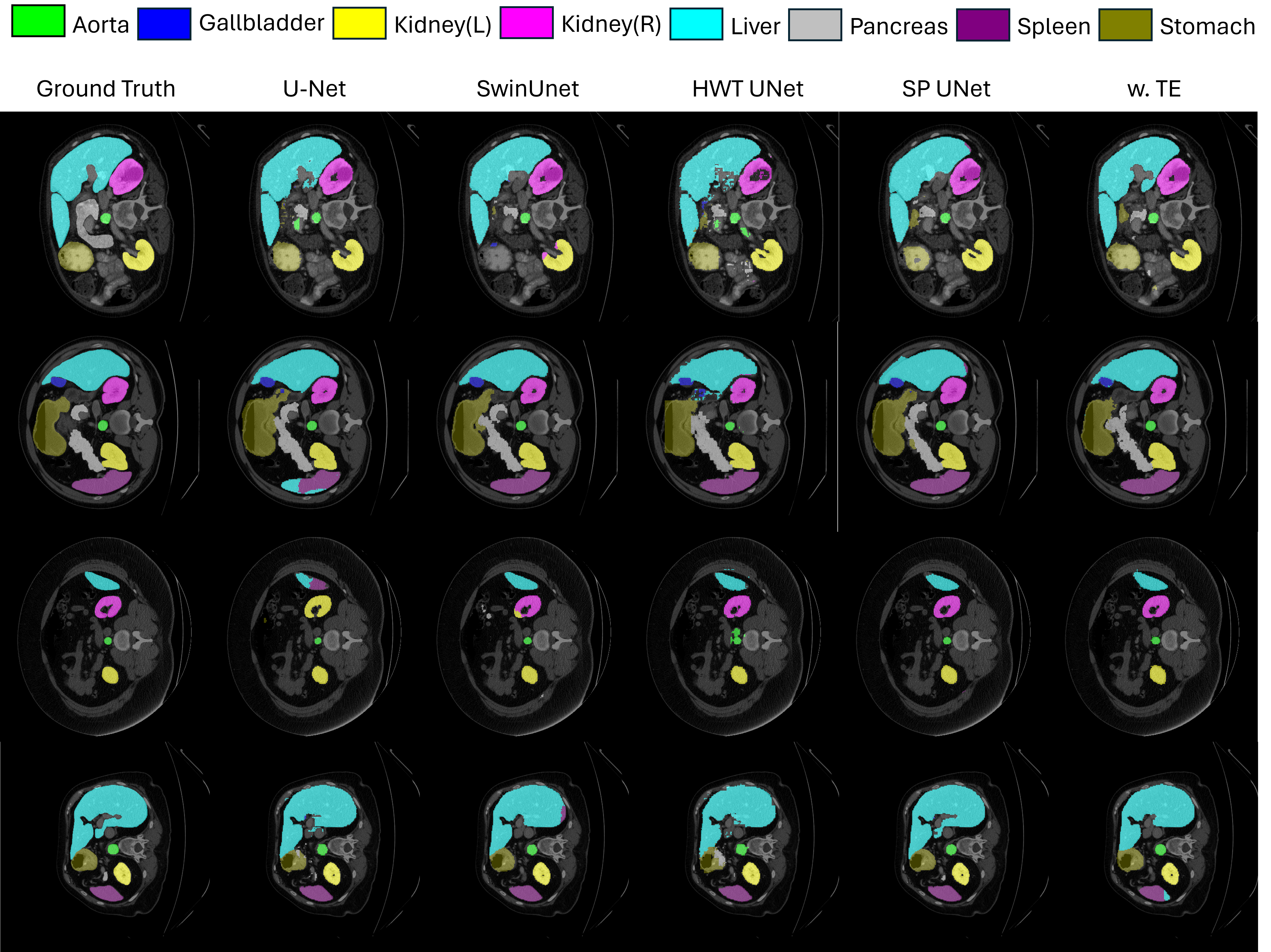}
    \caption{Qualitative comparison of different approaches on the Synapse dataset. From left to right: Ground Truth, U-Net, SwinUnet, UNet with HWT pooling, our SP UNet and the TE selected variant.}
    \label{fig:vis}
\end{figure}

\noindent\textbf{Comparison experiments on 2D benchmarks.} In this section, we present the segmentation accuracy of different methods on two 2D BIS benchmarks. Our proposed method, SP U-Net, integrates the U-Net architecture with our novel Stair Pooling approach. Additionally, we introduce a variant, denoted as "w. TE", which employs transfer entropy to search for optimal down-sampling paths, simplifying the network. Table.\ref{tab:synpase} shows the segmentation accuracy of various methods on the Synapse dataset. It is observed that Our proposed SP U-Net achieves the highest DSC of 80.45\% and a competitive HD of 31.25. It excels in segmenting the gallbladder (71.59\%) and pancreas (67.75\%) and shows strong performance across other organs. The variant with transfer entropy selection achieves the highest overall DSC of 80.89\% and an HD of 31.18. It shows the best performance in segmenting the aorta (90.2\%), left kidney (84.43\%), right kidney (79.76\%), and liver (95.09\%), with notable improvements in overall segmentation accuracy. Our methods outperform other pooling techniques, such as the Harr wavelet (HWT) and pyramid pooling techniques. Among other methods, SwinUnet stands out with an HD of 21.55 and a DSC of 79.13\%. However, its performance on individual organs, such as the pancreas (56.58\%) and spleen (90.66\%), is slightly lower compared to our proposed methods. A similar tendency is also observed in Table.\ref{tab:acdc} that presents the results on the ACDC dataset. Our SP U-Net achieves the highest DSC of 90.18\%, demonstrating its superior segmentation capability. It shows strong performance across all categories, with notable results in RV (88.81\%), Myo (84.63\%), and LV (95.90\%). The simpler variant also performs exceptionally well, with a DSC of 90.08\%. It shows the highest accuracy in LV segmentation (96.08\%) and competitive results in RV (87.74\%) and Myo (84.14\%).

\begin{wraptable}{r}{0.58\textwidth}
\centering
\caption{Segmentation accuracy of different methods on the KiT23 dataset in DSC($\%$).}\label{tab:kits}
\resizebox{0.58\textwidth}{!}{\begin{tabular}{lcccc}
\toprule
\multicolumn{1}{c}{Methods}           & DSC             & Kidney      & Masses      & Tumor        \\
\midrule
UNETER          & 73.4 & 95.54 & 69.07 & 60.56  \\
UNETER+SENet  & 74.9                 & 95.70 & 69.73 & 60.50  \\
UNETERt+CBAM   & 75.2                 & 95.33 & 70.05 & 62.46  \\
\rowcolor[rgb]{ .906,  .902,  .902} U-Net++           & 75.9                 & 95.65 & 70.94 & 62.64  \\
Attention
  UNETER & 76.6                 & 95.47 & 70.65 & 62.73  \\
  \midrule
SP UNETER          & \textbf{77.1}               & \textbf{95.71} & 71.77 & \textbf{63.85}  \\
w. TE             & 76.9                     & 95.46       & \textbf{71.93}       & 63.47    \\
\bottomrule
\end{tabular}}
\end{wraptable}
\noindent\textbf{Comparison experiments on 3D benchmark.} Table.\ref{tab:kits} presents the segmentation accuracy in terms of DSC for various methods on the KiT23 dataset. Our proposed SP UNETER achieves the highest overall DSC of 77.1\%, indicating its superior performance in 3D kidney tumor segmentation. It also shows the highest accuracy for kidney (95.71\%) and tumor (63.85\%) segmentation, and a strong performance in masses (71.77\%). It performs best in segmenting masses with a DSC of 71.93\% and shows competitive results for kidney (95.46\%) and tumor (63.47\%) segmentation.  The variant with searched optimal paths achieves an overall DSC of 76.9\%. It proves that our Stair Pooling and the transfer entropy selection can be easily extended to the 3D BIS task. The baseline 3D U-Net achieves an overall DSC of 73.4\%, with strong performance in kidney segmentation (95.54\%) but lower accuracy for masses (69.07\%) and tumor (60.56\%). Enhancing UNETER with SENet~\cite{hu2018squeeze} results in a slight improvement, reaching an overall DSC of 74.9\%. Attention UNETER achieves a DSC of 76.6\%, with strong performance in kidney and tumor segmentation, but is still slightly lower than our method.

Overall, these results on both 2D and 3D benchmarks demonstrate the effectiveness of our proposed SP U-Net and its simpler variant. These methods consistently achieve high segmentation accuracy, outperforming existing approaches in several key metrics. The integration of stair pooling proves to be beneficial in enhancing the segmentation performance of U-Net by significantly improving its efficacy in capturing long-range semantic information. \textbf{More encouragingly, although achieving similar overall DSC scores, SP U-Net and its TE variant perform quite differently across each category}. For instance, after applying TE selection, the segmentation performance of the right kidney (Kidney(R)) climbs from 76.53\% to 79.76\%, while the performance for the pancreas drops by 4.29\%. We speculate that this variation is due to the different shapes of organs or tissues. Utilizing a small and narrow pooling kernel ensures that important shape features are preserved and not directly pooled away. However, since our method calculates the entropy of the entire feature map, it may not be effective for every tissue or organ.

\begin{wrapfigure}{r}{0.5\textwidth}
  \centering
  \includegraphics[width=0.48\textwidth]{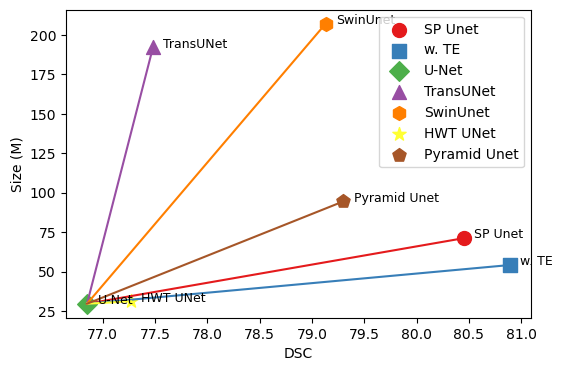}
  \caption{Segmentation results and sizes of different models on the Synapse dataset.}\label{fig:eff}
\end{wrapfigure}
\subsection{Qualitative Analysis} In Figure.\ref{fig:vis}, we provide qualitative comparison results on the Synapse dataset. This image illustrates the segmentation performance of different methods on medical images. The pure U-Net tends to over-segment and misclassify organs. For example, in the second row, the kidney (R) and liver are misclassified. Moreover, in the rest rows, the liver is also over-segmented. Compared to the Harr Wavelet Pooling U-Net (HWT UNet), our SP U-Net is better at encoding global contexts and distinguishing semantic features to improve the pure U-Net. We also observe that incorporating HWT pooling tends to generate noise in small regions. When comparing Transformer-based model, it can be observed that the predictions by SwinUnet tend to be coarser than those by SP U-Net regarding boundary and shape, especially the TE selected variant. For instance, in the first and second rows, the pancreas predictions by SwinUnet are less precise. Furthermore, in the third row, SP U-Net correctly segments both the left and right kidneys, while SwinUnet erroneously fills the inner hole of the right kidney. These observations suggest that SP U-Net excels in finer segmentation and preserving detailed shape information. The integration of stair pooling into the U-Net architecture allows SP U-Net to leverage long-range semantic interactions effectively. 

\subsection{Discussion}

\noindent\textbf{The efficiency of Stair Pooling.}\label{pa:eff} 
Figure.\ref{fig:eff} presents the segmentation performance
\begin{wraptable}{r}{0.6\textwidth}
\centering
\caption{Optimal paths selected by calculating TE on three datasets, where $P'_{vhw}$ refers to a $v\times h\times w$ pooling kernel.}\label{tab:path}
\resizebox{0.58\textwidth}{!}{\begin{tabular}{lccc} 
\toprule
Dataset & Step1          & Step2          & Step3           \\ 
\midrule
Synpase & $P'_{21}$,$P'_{12}$        & $P'_{21}$,$P'_{12}$        & $P'_{12}$,$P'_{21}$       \\
ACDC    & $P'_{21}$,$P'_{12}$        & $P'_{21}$,$P'_{12}$        & $P'_{21}$,$P'_{12}$         \\
KiT23   & $P'_{122}$,$P'_{212}$,$P'_{221}$ & $P'_{122}$,$P'_{221}$,$P'_{212}$ & $P'_{122}$,$P'_{212}$,$P'_{221}$  \\
\bottomrule
\end{tabular}}
\end{wraptable}
and model size for different neural network architectures on the Synapse dataset. The SP Unet model achieves a DSC of 80.45 with a moderate model size of 71.2M, indicating a strong performance. After TE selection, the variant boasts the highest DSC score of 80.89, and does so with a relatively lower model size of 54.2M, highlighting its efficiency. Our TE selection can help us further down-size the SP UNETER, from 143.6M to 65.8M. Notably, the sizes of the pure 2D and 3D U-Net models are 29.7M and 30.4M, respectively, which are slightly smaller than our TE variant. In contrast, the TransUNet model, while achieving a DSC of 77.48, has a significantly larger size of 192M. Similarly, the SwinUnet model has the largest model size at 207M. In our experiments, the pyramid pooling~\cite{he2015spatial} can boost the pure U-Net and bring an extra 63.8M. These experiments indicate that SP Unet and TE selection strategies provide a balanced trade-off between high segmentation performance and model size.

\begin{wrapfigure}{r}{0.4\textwidth}
  \centering
  \includegraphics[width=0.38\textwidth]{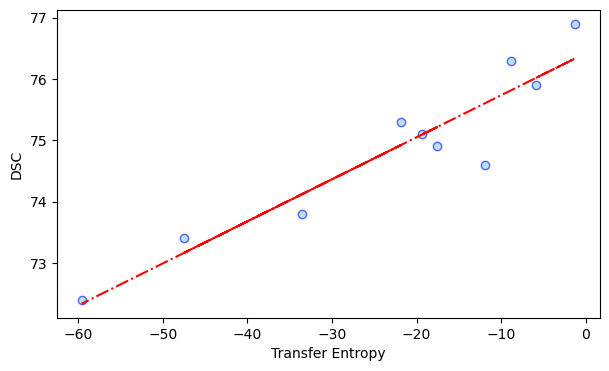}
  \caption{Correlations between DSC and TE on the KiT23 dataset.}\label{fig:fitline}
\end{wrapfigure}
\noindent\textbf{The optimal paths.} In Table \ref{tab:path}, we present the optimal paths in each downsampling step by leveraging the TE on three datasets. Our findings reveal that for the two 2D datasets, Synapse and ACDC, the models consistently prefer to perform horizontal pooling operations first. This preference suggests that beginning with horizontal pooling helps to retain more critical information, possibly due to the anatomical structures in these datasets being more distinguishable along the horizontal axis. In contrast, for the 3D dataset KiT23, the model's first pooling choice is along the z-axis of the voxels. This initial preference for z-axis pooling indicates that capturing depth information is crucial for effective segmentation in 3D medical images, given the complex spatial relationships present in volumetric data.

\begin{wrapfigure}{r}{0.4\textwidth}
  \centering
  \includegraphics[width=0.38\textwidth]{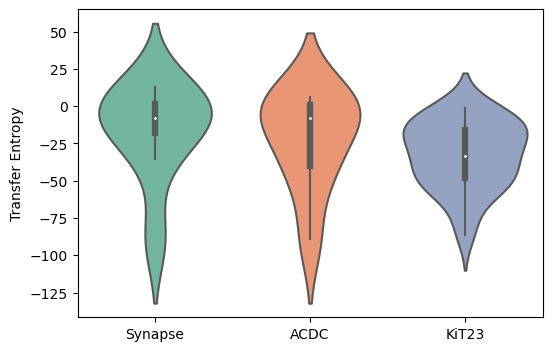}
  \caption{Violin plots for visualizing distributions of transfer entropy across different paths on three datasets.}\label{fig:volin}
\end{wrapfigure}
\noindent\textbf{Distribution of Transfer Entropy} Figure \ref{fig:volin} presents violin plots that visualize the distribution of transfer entropy (TE) across different variants on three datasets. For the Synapse dataset, the optimal variant achieves the highest TE value of 12.7, indicating a notable positive deviation. The average TE for this dataset is approximately -17.7, suggesting a general trend towards negative TE despite the presence of some high positive outliers. This indicates that not all the down-sampling paths effectively reduce information loss. The distribution on the ACDC dataset is similar, with the highest value being 5.9 and the average value around -22.4. However, in the KiT23 dataset, we observe a highest value of -1.3 and an average value of approximately -30.7. We further randomly selected nine variants with various down-sampling paths and evaluated their performances, presented in Figure.\ref{fig:fitline}. We found that the TE values calculated by the fused downsampling path feature $Z$ are generally proportional to the overall DSC. Therefore, we select the path with the highest TE as the optimal model.

\noindent\textbf{Limitation and Future Work} One of the limitations of our method is that for each downsampling step, we can only compute the feature of a single path, which is somewhat inadequate when dealing with 3D pooling. We found that the fused path contains the highest value of entropy in the KiT23 dataset. Therefore, our future work will focus on extending this method to automatically find the optimal path that includes the fused path, while also optimizing the search algorithm. 

\section{Conclusion}
In this paper, we introduced Stair Pooling, a novel down-sampling strategy designed to enhance the performance of U-Net architectures for biomedical image segmentation. Our method addresses the critical challenge of capturing long-range semantic interactions by utilizing small and narrow pooling layers in varied orientations to moderate the rate of down-sampling. This approach effectively preserves more critical information during the down-sampling process, balancing computational demands with the need for high information retention. Through extensive experiments on both 2D and 3D BIS benchmarks, we demonstrated that integrating Stair Pooling into the U-Net significantly improves their performance, achieving an average Dice score increase of 3.8\%. Additionally, by leveraging transfer entropy, we optimized the down-sampling paths, further simplifying the network without compromising performance. 

\newpage
\small{\bibliographystyle{abbrv}
\bibliography{egbib}}

@String(ICASSP=	{ICASSP})

@inproceedings{unet,
  title={U-net: Convolutional networks for biomedical image segmentation},
  author={Ronneberger, Olaf and Fischer, Philipp and Brox, Thomas},
  booktitle={MICCAI},
  pages={234--241},
  year={2015},
  organization={Springer}
}

@article{BISsurvey,
  title={Image segmentation using deep learning: A survey},
  author={Minaee, Shervin and Boykov, Yuri and Porikli, Fatih and Plaza, Antonio and Kehtarnavaz, Nasser and Terzopoulos, Demetri},
  journal={IEEE transactions on pattern analysis and machine intelligence},
  volume={44},
  number={7},
  pages={3523--3542},
  year={2021},
  publisher={IEEE}
}

@article{nnunet,
  title={nnU-Net: a self-configuring method for deep learning-based biomedical image segmentation},
  author={Isensee, Fabian and Jaeger, Paul F and Kohl, Simon AA and Petersen, Jens and Maier-Hein, Klaus H},
  journal={Nature methods},
  volume={18},
  number={2},
  pages={203--211},
  year={2021},
  publisher={Nature Publishing Group}
}

@inproceedings{soomro2018strided,
  title={Strided U-Net model: Retinal vessels segmentation using dice loss},
  author={Soomro, Toufique A and Afifi, Ahmed J and Gao, Junbin and Hellwich, Olaf and Paul, Manoranjan and Zheng, Lihong},
  booktitle={2018 Digital Image Computing: Techniques and Applications (DICTA)},
  pages={1--8},
  year={2018},
  organization={IEEE}
}

@inproceedings{riad2021learning,
  title={Learning Strides in Convolutional Neural Networks},
  author={Riad, Rachid and Teboul, Olivier and Grangier, David and Zeghidour, Neil},
  booktitle={International Conference on Learning Representations},
  year={2021}
}

@article{lecun1998gradient,
  title={Gradient-based learning applied to document recognition},
  author={LeCun, Yann and Bottou, L{\'e}on and Bengio, Yoshua and Haffner, Patrick},
  journal={Proceedings of the IEEE},
  volume={86},
  number={11},
  pages={2278--2324},
  year={1998},
  publisher={Ieee}
}

@inproceedings{huang2020unet3+,
  title={Unet 3+: A full-scale connected unet for medical image segmentation},
  author={Huang, Huimin and Lin, Lanfen and Tong, Ruofeng and Hu, Hongjie and Zhang, Qiaowei and Iwamoto, Yutaro and Han, Xianhua and Chen, Yen-Wei and Wu, Jian},
  booktitle={ICASSP 2020-2020 IEEE international conference on acoustics, speech and signal processing (ICASSP)},
  pages={1055--1059},
  year={2020},
  organization={IEEE}
}

@inproceedings{zhou2018unet++,
  title={Unet++: A nested u-net architecture for medical image segmentation},
  author={Zhou, Zongwei and Rahman Siddiquee, Md Mahfuzur and Tajbakhsh, Nima and Liang, Jianming},
  booktitle={Deep Learning in Medical Image Analysis and Multimodal Learning for Clinical Decision Support: 4th International Workshop, DLMIA 2018, and 8th International Workshop, ML-CDS 2018, Held in Conjunction with MICCAI 2018, Granada, Spain, September 20, 2018, Proceedings 4},
  pages={3--11},
  year={2018},
  organization={Springer}
}

@inproceedings{cao2022swinu,
  title={Swin-unet: Unet-like pure transformer for medical image segmentation},
  author={Cao, Hu and Wang, Yueyue and Chen, Joy and Jiang, Dongsheng and Zhang, Xiaopeng and Tian, Qi and Wang, Manning},
  booktitle={European conference on computer vision},
  pages={205--218},
  year={2022},
  organization={Springer}
}

@article{chen2021transunet,
  title={Transunet: Transformers make strong encoders for medical image segmentation},
  author={Chen, Jieneng and Lu, Yongyi and Yu, Qihang and Luo, Xiangde and Adeli, Ehsan and Wang, Yan and Lu, Le and Yuille, Alan L and Zhou, Yuyin},
  journal={arXiv preprint arXiv:2102.04306},
  year={2021}
}

@article{wang2024mamba,
  title={Mamba-unet: Unet-like pure visual mamba for medical image segmentation},
  author={Wang, Ziyang and Zheng, Jian-Qing and Zhang, Yichi and Cui, Ge and Li, Lei},
  journal={arXiv preprint arXiv:2402.05079},
  year={2024}
}

@inproceedings{cciccek20163d,
  title={3D U-Net: learning dense volumetric segmentation from sparse annotation},
  author={{\c{C}}i{\c{c}}ek, {\"O}zg{\"u}n and Abdulkadir, Ahmed and Lienkamp, Soeren S and Brox, Thomas and Ronneberger, Olaf},
  booktitle={Medical Image Computing and Computer-Assisted Intervention--MICCAI 2016: 19th International Conference, Athens, Greece, October 17-21, 2016, Proceedings, Part II 19},
  pages={424--432},
  year={2016},
  organization={Springer}
}

@article{ma2024segment,
  title={Segment anything in medical images},
  author={Ma, Jun and He, Yuting and Li, Feifei and Han, Lin and You, Chenyu and Wang, Bo},
  journal={Nature Communications},
  volume={15},
  number={1},
  pages={654},
  year={2024},
  publisher={Nature Publishing Group UK London}
}

@inproceedings{zhao2020liftpool,
  title={LiftPool: Bidirectional ConvNet Pooling},
  author={Zhao, Jiaojiao and Snoek, Cees GM},
  booktitle={International Conference on Learning Representations},
  year={2020}
}

@inproceedings{zhou2023xnet,
  title={Xnet: Wavelet-based low and high frequency fusion networks for fully-and semi-supervised semantic segmentation of biomedical images},
  author={Zhou, Yanfeng and Huang, Jiaxing and Wang, Chenlong and Song, Le and Yang, Ge},
  booktitle={Proceedings of the IEEE/CVF International Conference on Computer Vision},
  pages={21085--21096},
  year={2023}
}

@article{xu2023haar,
  title={Haar wavelet downsampling: A simple but effective downsampling module for semantic segmentation},
  author={Xu, Guoping and Liao, Wentao and Zhang, Xuan and Li, Chang and He, Xinwei and Wu, Xinglong},
  journal={Pattern Recognition},
  volume={143},
  pages={109819},
  year={2023},
  publisher={Elsevier}
}

@article{lin2024mlp,
  title={MLP Can Be A Good Transformer Learner},
  author={Lin, Sihao and Lyu, Pumeng and Liu, Dongrui and Tang, Tao and Liang, Xiaodan and Song, Andy and Chang, Xiaojun},
  journal={arXiv preprint arXiv:2404.05657},
  year={2024}
}

@article{sun2022entropy,
  title={Entropy-driven mixed-precision quantization for deep network design},
  author={Sun, Zhenhong and Ge, Ce and Wang, Junyan and Lin, Ming and Chen, Hesen and Li, Hao and Sun, Xiuyu},
  journal={Advances in Neural Information Processing Systems},
  volume={35},
  pages={21508--21520},
  year={2022}
}

@inproceedings{hu2018squeeze,
  title={Squeeze-and-excitation networks},
  author={Hu, Jie and Shen, Li and Sun, Gang},
  booktitle={Proceedings of the IEEE conference on computer vision and pattern recognition},
  pages={7132--7141},
  year={2018}
}

@article{zhao2023wranet,
  title={WRANet: wavelet integrated residual attention U-Net network for medical image segmentation},
  author={Zhao, Yawu and Wang, Shudong and Zhang, Yulin and Qiao, Sibo and Zhang, Mufei},
  journal={Complex \& intelligent systems},
  volume={9},
  number={6},
  pages={6971--6983},
  year={2023},
  publisher={Springer}
}

@article{nirthika2022pooling,
  title={Pooling in convolutional neural networks for medical image analysis: a survey and an empirical study},
  author={Nirthika, Rajendran and Manivannan, Siyamalan and Ramanan, Amirthalingam and Wang, Ruixuan},
  journal={Neural Computing and Applications},
  volume={34},
  number={7},
  pages={5321--5347},
  year={2022},
  publisher={Springer}
}

@article{du2021unet,
  title={UNet retinal blood vessel segmentation algorithm based on improved pyramid pooling method and attention mechanism},
  author={Du, Xin-Feng and Wang, Jie-Sheng and Sun, Wei-zhen},
  journal={Physics in Medicine \& Biology},
  volume={66},
  number={17},
  pages={175013},
  year={2021},
  publisher={IOP Publishing}
}

@inproceedings{hatamizadeh2022unetr,
  title={Unetr: Transformers for 3d medical image segmentation},
  author={Hatamizadeh, Ali and Tang, Yucheng and Nath, Vishwesh and Yang, Dong and Myronenko, Andriy and Landman, Bennett and Roth, Holger R and Xu, Daguang},
  booktitle={Proceedings of the IEEE/CVF winter conference on applications of computer vision},
  pages={574--584},
  year={2022}
}

@article{li2023automatic,
  title={Automatic segmentation of prostate MRI based on 3D pyramid pooling Unet},
  author={Li, Yuchun and Lin, Cong and Zhang, Yu and Feng, Siling and Huang, Mengxing and Bai, Zhiming},
  journal={Medical Physics},
  volume={50},
  number={2},
  pages={906--921},
  year={2023},
  publisher={Wiley Online Library}
}

@inproceedings{liu2021application,
  title={Application of hybrid network of UNet and feature pyramid network in spine segmentation},
  author={Liu, Xingxing and Deng, Wenxiang and Liu, Yang},
  booktitle={2021 IEEE International Symposium on Medical Measurements and Applications (MeMeA)},
  pages={1--6},
  year={2021},
  organization={IEEE}
}

@inproceedings{oktay2022attention,
  title={Attention U-Net: Learning Where to Look for the Pancreas},
  author={Oktay, Ozan and Schlemper, Jo and Le Folgoc, Loic and Lee, Matthew and Heinrich, Mattias and Misawa, Kazunari and Mori, Kensaku and McDonagh, Steven and Hammerla, Nils Y and Kainz, Bernhard and others},
  booktitle={Medical Imaging with Deep Learning},
  year={2022}
}

@article{he2015spatial,
  title={Spatial pyramid pooling in deep convolutional networks for visual recognition},
  author={He, Kaiming and Zhang, Xiangyu and Ren, Shaoqing and Sun, Jian},
  journal={IEEE transactions on pattern analysis and machine intelligence},
  volume={37},
  number={9},
  pages={1904--1916},
  year={2015},
  publisher={IEEE}
}

@inproceedings{alsallakh2022mind,
  title={Mind the Pool: Convolutional Neural Networks Can Overfit Input Size},
  author={Alsallakh, Bilal and Yan, David and Kokhlikyan, Narine and Miglani, Vivek and Reblitz-Richardson, Orion and Bhattacharya, Pamela},
  booktitle={The Eleventh International Conference on Learning Representations},
  year={2022}
}

@article{stergiou2022adapool,
  title={Adapool: Exponential adaptive pooling for information-retaining downsampling},
  author={Stergiou, Alexandros and Poppe, Ronald},
  journal={IEEE Transactions on Image Processing},
  volume={32},
  pages={251--266},
  year={2022},
  publisher={IEEE}
}

@inproceedings{hou2020strip,
  title={Strip pooling: Rethinking spatial pooling for scene parsing},
  author={Hou, Qibin and Zhang, Li and Cheng, Ming-Ming and Feng, Jiashi},
  booktitle={Proceedings of the IEEE/CVF conference on computer vision and pattern recognition},
  pages={4003--4012},
  year={2020}
}

@inproceedings{zhai2017s3pool,
  title={S3pool: Pooling with stochastic spatial sampling},
  author={Zhai, Shuangfei and Wu, Hui and Kumar, Abhishek and Cheng, Yu and Lu, Yongxi and Zhang, Zhongfei and Feris, Rogerio},
  booktitle={Proceedings of the IEEE conference on computer vision and pattern recognition},
  pages={4970--4978},
  year={2017}
}

@inproceedings{saeedan2018detail,
  title={Detail-preserving pooling in deep networks},
  author={Saeedan, Faraz and Weber, Nicolas and Goesele, Michael and Roth, Stefan},
  booktitle={Proceedings of the IEEE Conference on Computer Vision and Pattern Recognition},
  pages={9108--9116},
  year={2018}
}

@inproceedings{gao2019lip,
  title={Lip: Local importance-based pooling},
  author={Gao, Ziteng and Wang, Limin and Wu, Gangshan},
  booktitle={Proceedings of the IEEE/CVF International Conference on Computer Vision},
  pages={3355--3364},
  year={2019}
}

@inproceedings{stergiou2021refining,
  title={Refining activation downsampling with SoftPool},
  author={Stergiou, Alexandros and Poppe, Ronald and Kalliatakis, Grigorios},
  booktitle={Proceedings of the IEEE/CVF international conference on computer vision},
  pages={10357--10366},
  year={2021}
}

@inproceedings{li2023dynamic,
  title={Dynamic graph enhanced contrastive learning for chest x-ray report generation},
  author={Li, Mingjie and Lin, Bingqian and Chen, Zicong and Lin, Haokun and Liang, Xiaodan and Chang, Xiaojun},
  booktitle={Proceedings of the IEEE/CVF conference on computer vision and pattern recognition},
  pages={3334--3343},
  year={2023}
}

@article{islamaj2024overview,
  title={The overview of the BioRED (biomedical relation extraction dataset) track at BioCreative VIII},
  author={Islamaj, Rezarta and Lai, Po-Ting and Wei, Chih-Hsuan and Luo, Ling and Almeida, Tiago and Jonker, Richard A A and Concei{\c{c}}{\~a}o, Sofia I R and Sousa, Diana F and Phan, Cong-Phuoc and Chiang, Jung-Hsien and others},
  journal={Database},
  volume={2024},
  pages={baae069},
  year={2024},
  publisher={Oxford University Press UK}
}

@article{chen2025inference,
  title={Inference-specific learning for improved medical image segmentation},
  author={Chen, Yizheng and Liu, Sheng and Li, Mingjie and Han, Bin and Xing, Lei},
  journal={Medical Physics},
  volume={52},
  number={7},
  pages={e17883},
  year={2025},
  publisher={Wiley Online Library}
}

@article{li2022mitigating,
  title={Mitigating Data Redundancy to Revitalize Transformer-based Long-Term Time Series Forecasting System},
  author={Li, Mingjie and Liu, Rui and Shi, Guangsi and Han, Mingfei and Li, Changlin and Yao, Lina and Chang, Xiaojun and Chen, Ling},
  journal={ACM Transactions on Intelligent Systems and Technology},
  publisher={ACM New York, NY},
  year={2025}
}

@article{weng2023mask,
  title={Mask propagation for efficient video semantic segmentation},
  author={Weng, Yuetian and Han, Mingfei and He, Haoyu and Li, Mingjie and Yao, Lina and Chang, Xiaojun and Zhuang, Bohan},
  journal={Advances in Neural Information Processing Systems},
  volume={36},
  pages={7170--7183},
  year={2023}
}

\end{document}